## Title

SpineWave: Harnessing Fish Rigid-Flexible Spinal Kinematics for Enhancing Biomimetic Robotic Locomotion

## Authors

Qu He[1,3†], Weikun Li[3,4†,*], Guangmin Dai[1,3†], Hao Chen[3,4†], Qimeng Liu[1,3], Xiaoqing Tian[5], Jie You[6], Weicheng Cui[2,3,4*], Michael S. Triantafyllou[7,8], Dixia Fan[2,3,4*]

## Affiliations

[1] Zhejiang University, Hangzhou, Zhejiang 310027, China.
[2] Research Center for Industries of the Future, Westlake University, Hangzhou, Zhejiang 310030, China.
[3] Key Laboratory of Coastal Environment and Resources of Zhejiang Province, School of Engineering, Westlake University, Hangzhou, Zhejiang 310030, China.
[4] Institute of Advanced Technology, Westlake Institute for Advanced Study, Hangzhou, Zhejiang 310024, China.
[5] Hangzhou Dianzi University, Hangzhou, Zhejiang 310018, China.
[6] School of Ocean Sciences, China University of Geosciences, Beijing 100083, China
[7] Department of Mechanical Engineering, Massachusetts Institute of Technology, Cambridge, MA 02139, USA.
[8] MIT Sea Grant College Program, Massachusetts Institute of Technology, Cambridge, MA 02139, USA.

*Corresponding author. Email: fandixia@westlake.edu.cn, cuiweicheng@westlake.edu.cn, liweikun@westlake.edu.cn

[†] These authors contributed equally to this work.

## Abstract

Fish have endured millions of years of evolution, and their distinct rigid-flexible body structures offer inspiration for overcoming challenges in underwater robotics, such as limited mobility, high energy consumption, and adaptability. This paper introduces SpineWave, a biomimetic robotic fish featuring a fish-spine-like rigid-flexible transition structure. The structure integrates expandable fishbone-like ribs and adjustable magnets, mimicking the stretch and recoil of fish muscles to balance rigidity and flexibility. In addition, we employed an evolutionary algorithm to optimize robot's hydrodynamics, achieving significant improvements in swimming performance. Real-world tests demonstrated the robot's robustness and potential for environmental monitoring, underwater exploration, and industrial inspection, establishing SpineWave as a transformative platform for aquatic robotics.

**Introduction**

Over recent decades, the growing demand for underwater exploration and resource utilization has driven significant advancements in underwater robotic technologies[1,2], helping to address global challenges, including the sustainable use of ocean resources and the preservation of fragile marine ecosystems. While traditional remotely operated vehicles (ROVs) and autonomous underwater vehicles (AUVs)[3] address some needs, they face inherent limitations. ROVs are constrained by tethers[4], and AUVs often suffer from inefficient propulsion and low adaptability in complex environments[5]. A key challenge lies in replicating the remarkable locomotion efficiency and agility observed in fish, which stems from a sophisticated combination of rigid skeletal structures and flexible musculature. This natural synergy, honed over millions of years of evolution[6], allows fish to achieve steady swimming, swift reactions, and agile maneuvers—capabilities crucial for navigating complex aquatic environments and currently lacking in robotic counterparts. Addressing this gap by effectively integrating rigidity and flexibility is central to advancing biomimetic underwater robotics.

Ancient fish species, such as Haikouichthys and Myllokunmingia, emerging around 530 million years ago. likely descendants of lancelets, small worm-like invertebrates[7]. These early fish developed a basic vertebrate body plan, including a notochord, rudimentary vertebrae, head and tail[8,9], leading to robust skeletal structures and supple musculature[10] that enhanced maneuverability[11]. These features have been carried on to the modern fish, resulting in a compliant body with a rigid spine, flexible muscles, and a powerful tail. This biomechanical synergy[12] enables steady swimming, swift reactions, and agile maneuvers—crucial for evading predators and hunting prey. This type of locomotive abilities, especially in complex aquatic environments, has long attracted the attention of researchers[13,14], and inspired different underwater robotic design[15].

Fish-inspired robot development, highlighted by MIT's RoboTuna, spurred continued field advancements[16]. Robots like SPC-III[17] and MAR[18] use undulating propulsion for maneuverability and stability. Amphibious robots like Salamandra Robotica-II[19] combine undulation for swimming with coordinated limb motion for land walking. Fast fish-inspired robots like iSplash[20,21] use carangiform swimming for high speeds. Eco-friendly robots like Envirobot[22] use undulation-based propulsion instead of propellers, minimizing noise and ecological impact. Underactuated robotic fish also gained attention for simple designs and easy implementation, with promising results[23,24]. Magnetic actuation and innovative methods for biomimetic robotic fish with advanced transmission systems enhanced joint flexibility[25–27]. Improved Central Pattern Generator (CPG) models led to biomimetic robotic fish with CPG-based control systems, further advancing the field[28–30]. These innovations show increasing efficiency, ecological compatibility, and adaptability of bio-inspired robotic designs.

While promising, fish-inspired underwater robots face challenges[49]. Current designs are often either completely rigid or entirely soft (Fig. 1 B, D). The first type uses a rigid structure driven by servo motors, which has shown success in structural solidity in controlled environments[31,32]. The second category focuses on soft robotic fish, typically made from flexible materials like silicone[33] and powered by muscle-mimicking pneumatic[34] or pressure-resilient devices[35]. Notable examples, such as SoFi[36], demonstrate progress in integration and miniaturization. However, fish achieve efficient, agile underwater locomotion through a synergy of rigid skeletal structure and soft tissue (Fig. 1 C). This inspired our prototype, aiming to bridge this gap by integrating rigidity and flexibility, mimicking fish mechanics. Our research pioneers a biomimetic fish robot combining these features novelly.

This paper presents SpineWave, a new design paradigm from fish anatomy. By blending rigidity (emulating vertebrae with metal/polymer joints) and passive flexibility (mimicking muscles with magnetism for stretching/rebounding), we follow the "derived from nature, returning to nature" principle. Prior art used magnetic actuation via ribcages for traveling-wave propulsion[27]; our design uses a rigid-flexible spine with passive magnets assisting servo motors for a simpler, lower-cost design optimized with EGO and Kriging modeling. We aim to provide a standardized, modular, scalable platform (Fig. 2) for researchers, facilitating rapid prototyping and experimentation across fish morphologies (tuna to eel) and serving as a robust optimization base.

We embraced a modular, scalable design for "SpineWave," a comprehensive bionic fish robot platform. Standardized skeletal components allow quick modifications, enabling transitions between fish forms (tuna to coelacanth to eel). The platform is an efficient, stable base for optimization. Using evolutionary algorithms, we optimized propulsion and steering, enhancing stability and underscoring its potential for continuous improvement. Our biomimetic robotic fish achieves life-like swimming, similar to real fish. (Supplementary Movie S1)

We also conducted rigorous testing in natural aquatic settings (long-distance swims, complex terrains, salt water) further demonstrating SpineWave's enhanced performance and adaptability in real-world environments.

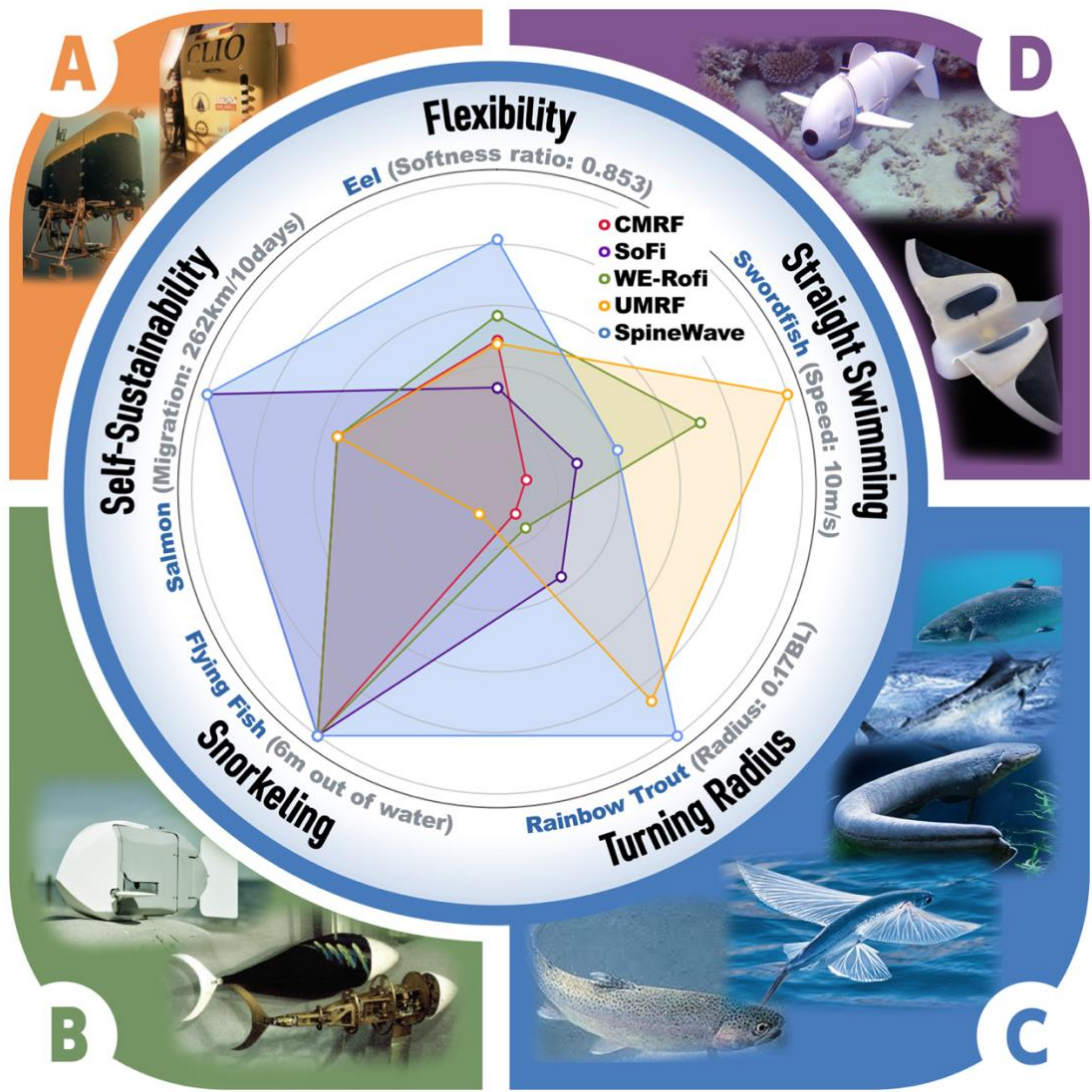

**Fig.1 Graduation of body rigidity swimming performance in aquatic locomotion.** The counterclockwise directions in the figure are **(A)** rigid autonomous underwater vehicles (AUV)[37,38], **(B)** rigid, articulated, or pulley robotic fish[21,39], **(C)** real fish with soft and flexible body parts[40], **(D)** completely soft robotic fish[35,36], and **(Center)** the swimming performance (Including five key metrics: Speed, Turning Radius, Flexibility, measured as the ratio of flexible body parts; Snorkeling, whether the robot fish can surface for respiration; Self-Sustainability, autonomous operation from controlled labs to real environments, detailed definitions are provided in the Supplementary Materials Note S1) of SpineWave prototype are compared to other novel fish robots such as SoFi[36], CMRF[41], WE-Rofi[42] and UMRF[43] (More detailed comparison can be found in the Supplementary Materials Table S1) and also real-world fishes[44–48] (We have listed the real fish with excellent performance in each performance metric). Spinwaves mainly draws inspiration from C, but ultilizes design philosophies from all 4 groups.

## Results

### Robotic Design and Construction

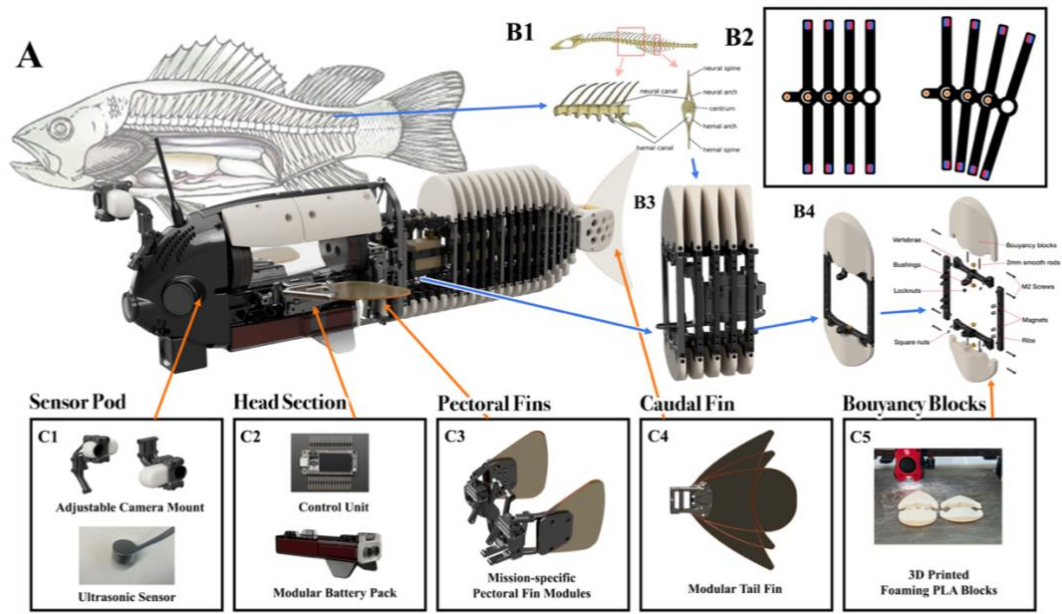

**Fig.2 Design and Assembly of the *SpineWave*. (A) The 3D-rendered model** showcases interchangeable functional modules, each replicating aspects of a biological fish's anatomy. The modular design ensures flexibility and adaptability in construction and functionality. **(B1 and B3) The hard endoskeleton module** is incorporated with five ribcage units along with a servo motor. This internal endoskeleton draws inspiration from a fish's rigid bony structure, which provides essential protection and enables agile locomotion. **(B2) Magnetic ribcages** are the key components of our magnetic exoskeleton structure. Each ribcage houses eight magnets arranged so that opposing magnets generate repulsive forces. **(B4) The magnetic exoskeleton** is the core component of the Ribcage Unit, and the exploded view illustrates its constituent parts. The exoskeleton emulates a fish's soft muscular structure. **(C1) The sensor Module** is located at the front and is equipped with two ultrasonic distance sensors and two cameras. **(C2) Electronics Pod and Battery** innovatively introduce the concept and technology of engineering modularization, considering both high reliability and easy maintenance. **(C3) Removable Pectoral Fin Modules** can be swapped between missions for different swimming modes. **(C4) Tail Assembly** with **Modular Caudal Fins** has a quick-swap attachment design, facilitating experimentation with caudal fins of varying shapes and materials. **(C5) Buoyancy** is made from 3D-printed foaming PLA blocks with a rounded, hydrodynamic shape akin to fish fat tissues.

SpineWave employs a fully modular architecture comprising interchangeable head, body, and tail units linked by waterproof connectors that enable rapid reconfiguration and straightforward maintenance. The streamlined head shell encloses the electronics pod, battery, sensor array and detachable servo-driven pectoral fins for enhanced maneuverability. The central body integrates five actively actuated internal segments with a passive magnetic exoskeleton and buoyancy blocks to mimic fish musculature and natural flotation. A screw-mounted interchangeable tail fin delivers thrust, with profiles easily swapped to suit mission requirements. Teleoperation is achieved via 433 MHz radio and 5.8 GHz camera for video streaming, and depth is controlled through adjustable buoyancy and ballast. In endurance trials, a three-segment SpineWave swam 800 m in 50 min.

All fabrication details, component specifications, magnetic-spring theory, control electronics, communication hardware and additional performance data are provided in Supplementary Material S9.

**Magnetically constrained bionic exoskeleton**

To assess the magnets' effect on bending, we compared rigidly constrained (simulating traditional servo fish robots) and magnetic segments (our design) (Fig. 3A). Constrained joints reach 30° per segment (red line), unconstrained reach 50° per segment (blue line), showing the latter's greater flexibility and maneuverability. Two kinematic simulation programs were developed: one models joint angle vs. magnetic repulsion (using magnet parameters and frame dimensions); the other uses simulated magnetic force and endoskeleton servo angles to find exoskeleton joint angles. Simulations closely match experimental results (Fig. 3B). (Comparisons in Supplementary Material Note S2, Fig.S2 and Movie S2)

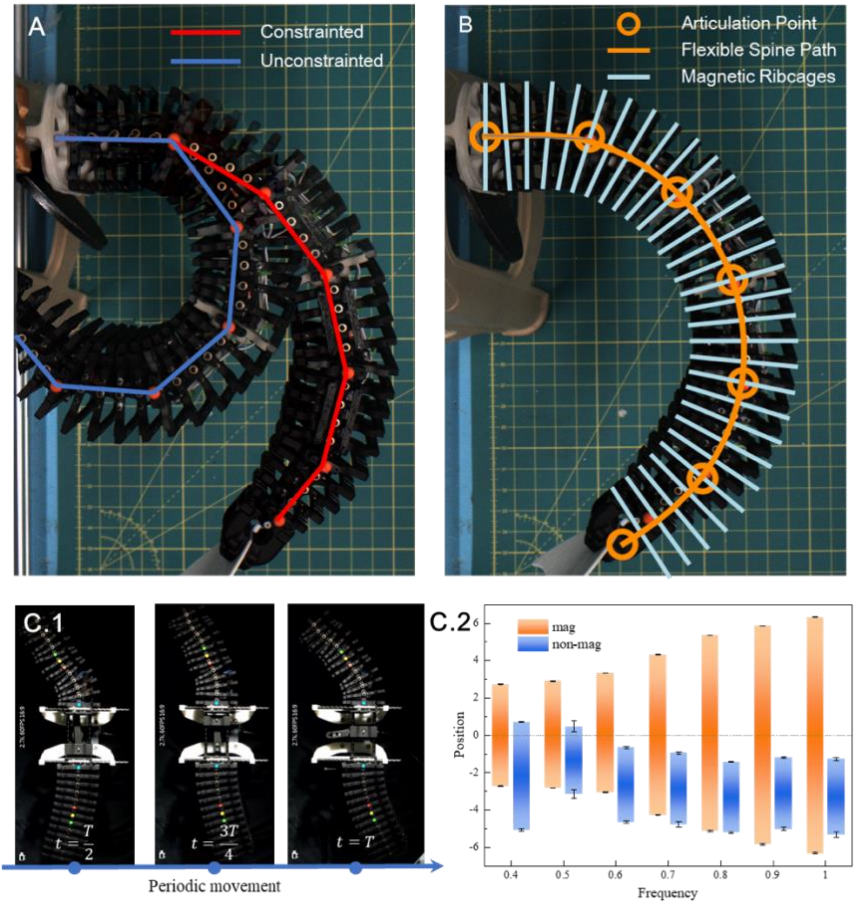

**Fig.3** (A) Comparison of Maximum Bending Capability: Bending performance of segments with constrained (red) and unconstrained (blue) magnetic joints. Constrained segments (with a rigid constraint bar) achieved a maximum bending angle of 30° per joint, while unconstrained segments reached 50°. (B) Experimental vs. Theoretical Bending Angles: The flexible spine (orange line) formed by magnetic joints and rib-like segments shows experimental angles matching theoretical model predictions, validating the magnetic skeleton design. (C.1) Kinematic Deformation Comparison: Exoskeleton module with (left) and without (right) magnetic constraints at $t = T/2$, $3T/4$, $T$, frequency $f$ = 0.7 Hz. Joints 1, 6, and 7 are marked for motion tracking. (C.2) Amplitude Comparison at Joint 7: Red (with magnetic constraints) vs. blue (without). Without constraints, the module shows asymmetrical movement and larger amplitude variation across cycles.

Due to the challenging nature of full dynamic simulation, our analysis focused on demonstrating magnetic constraints' stability enhancement over non-magnetic systems. While predicting passive joint angles in water is a future goal, this validates stable body wave maintenance. Experimentally, we compared 13-rib exoskeleton modules with/without magnets (replacing magnets with steel balls for equal weight). (Device details: Supplementary Material Note S3). Fig. 3C.1 shows module deformation at different cycle instants (t = T/2,3T/4,T) at f=0.7Hz. The magnetic module showed smooth, waveform-conforming movement; the non-magnetic module showed chaotic motion.

Fig. 3C.2 compares joint 7 amplitude at different frequencies. The magnetic module was stable around baseline (y=0); the non-magnetic showed larger variations with non-zero mean offsets. Simulations confirmed these results. (Details: Supplementary Material Note S4, Supplementary Movie S3).

**Scalable modules for emulating diverse BCF fish locomotion**

SpineWave mimics fish skeletal structure[55,56] with a modular, scalable design of interconnected, vertebra-like jointed parts. These offer support and undulatory movement flexibility. The modular approach allows quick customization for different fish species' swimming. Each segment is independently balanced and tuned for buoyancy, so segment number changes don't affect overall weight/buoyancy. Current modularity is limited by control system complexity; future work may explore more, but current design (1, 3, 5 servo segments) balances performance and control.

To show hardware modularity, three biomimetic robotic fish were built (Fig. 4), serving as testbeds for thunniform, subcarangiform, and anguilliform locomotion (Fig. 4). (Details/videos: Supplementary Materials Note S5, Supplementary Movie S4-S6).

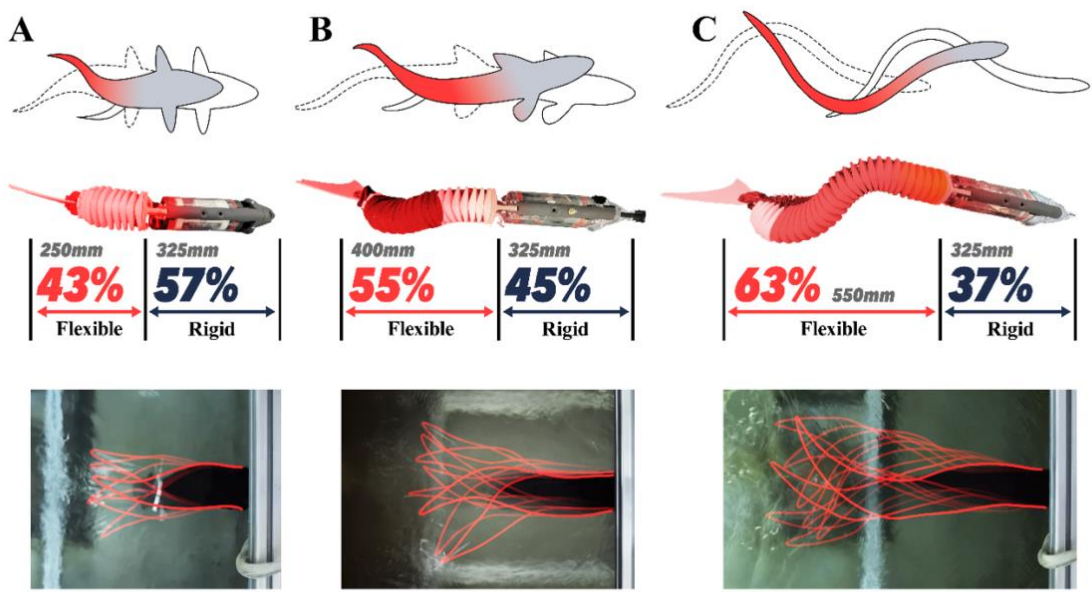

**Fig.4 Comparison of three fish robot prototypes. (From Top to Bottom)** Simplified diagram of real fish. Body proportions of SpineWave-I/II/III. Top view of three SpineWave prototypes on the testbench, image traced and stacked to showcase flexible section of the body. **Column (A)** Thuniform **(B)** Subcarangiform **(C)** Anguilliform.

**Evolutionary optimization for hydrodynamics**

To enhance SpineWave's swimming performance, we first optimized its hydrodynamics in a stationary configuration (Fig. 5B) and then applied the optimized control strategy to a free-swimming version (Fig. 5C). A Central Pattern Generator (CPG)[57] model controlled robot movement using seven parameters that were optimized by an in-house Efficient Global Optimization (EGO) algorithm[58]. Objectives varied by flow conditions: for straight-line swimming (Fig. 5B1), we aimed to maximize positive mean thrust; for side flow conditions (Fig. 5B2), we minimized mean thrust and torque for stable orientation; in turbulent environments (Fig. 5B3), we sought a swimming pattern with near-zero mean thrust behind an obstacle, utilizing vortex shedding for energy savings[59,60]; and for turning in still water (Fig. 5B4), we optimized for maximum torque about the center of gravity. In the experiment, each robot was connected to a six-axis force sensor, and a CCD camera placed above the tank captured its motion. For additional experimental details, please refer to Supplementary Materials Note S6 and Movie S7. A brief overview of the CPG model and EGO algorithm is provided in the Materials and Methods section.

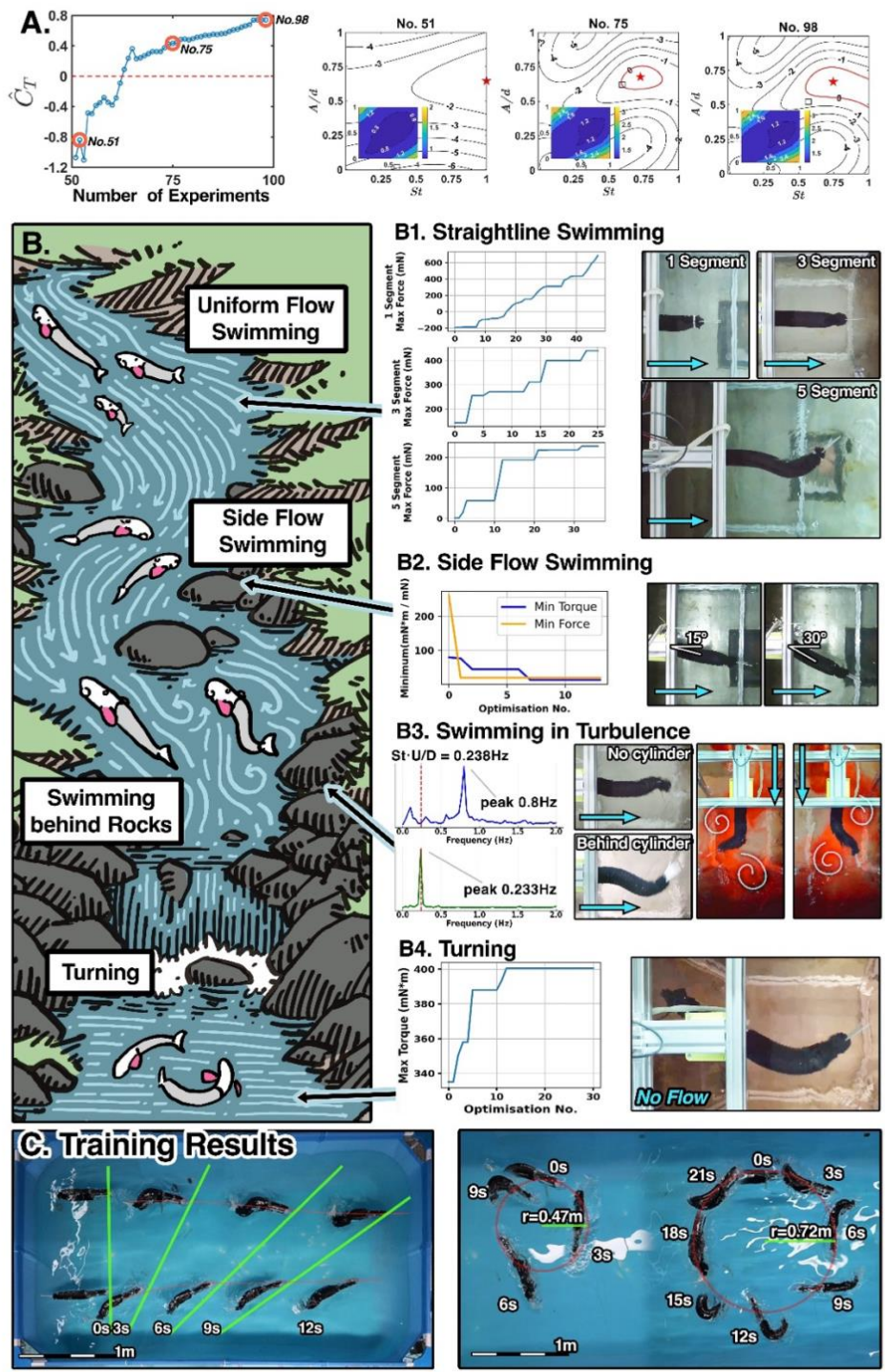

**Fig.5** (A) Hydrodynamic Optimization Process: Active learning sequence optimizing the mean thrust coefficient ($\hat{C}_T$) of SpineWave in uniform flow ($U$ = 0.3 m/s). The graph shows $\hat{C}_T$ progression across experiments, highlighting pre-optimization (No. 51), during optimization (No. 75), and optimal point (No. 98). Contour plots display predicted $\hat{C}_T$ as a function of Strouhal number ($St$) and tail amplitude ($A/d$), with insets showing prediction standard deviation (SD). Black dots: existing data; black squares: new experiments; red contour: $\hat{C}_T$ = 0; red stars: maximum $\hat{C}_T$. (B top to bottom) Experimental training scenarios adapt SpineWave to various flow conditions, including SpineWave-I/II/III in swimming uniform flow, SpineWave-II in side flow, SpineWave-II harnessing vortex behind obstacles, and SpineWave-II turning: (B1) Maximum thrust in constant flow, with increasing force during training. (B2) Near-zero torque/thrust in tilted flow (15°/30°), enabling stable orientation. (B3) Energy-saving behavior (29% reduction) behind a cylinder, simulating vortex use. Left: FFT shows frequency matching; right: SpineWave navigating vortices. (B4) Maximum torque optimization for enhanced turning performance. (C) Performance comparison before/after optimization, showing significant improvements in swimming efficiency and maneuverability.

Fig. 5A illustrates a typical EGO process and results of the robot in the uniform flow at $U = 0.3 m/s$. We plot the sequences of $C_T$ versus Strouhal number $St$ and non-dimensional tail amplitude $A/d$, with predicted SD contour inset. In addition, we highlight the zero-contour line ($\hat{C}_T = 0$) of the thrust and drag transition, showing clearly the discovery of thrust production region. In addition, it is worth noting that between experiment number 91[th] and 98[th], the $\hat{C}_T$ contours show only slight changes, indicating the convergence. For more specific details, please refer to Supplementary Material Note S7.

**Close water swimming performance**

To evaluate the robot's performance after optimization, we conducted straight-line swimming and turning experiments in a 4m × 2m water tank. Fig. 5C shows top-view videos and motor actions before and after optimization for both swimming (Fig. 5C-Left) and turning (Fig. 5C-Right). Pre-optimized parameters were selected from 50 random samples in the initial Kriging model with the best hydrodynamic performance. Significant improvements were seen, especially in swimming speed, turning velocity, and radius. The head's lateral motion was reduced, indicating improved stability.

Normalized by body length (BL: 0.725m), the straight-line swimming speed increased from 0.32 BL/s to 0.44 BL/s (38% improvement). After optimization, the robot completed a 360-degree turn in 10 seconds, down from 23 seconds, with a turning radius reduced by 35% (from 1BL to 0.65BL). These improvements demonstrate enhanced performance, supporting the robot's capabilities in complex environments. For more details, see Supplementary Movies S8-S9.

**Open water human-robot interaction**

After optimization, we first tested the SpineWave in open water at Westlake University's Yungu campus (Fig. 6A). The 3-joint SpineWave featured a dual camera system, including an underwater drone camera and an Insta360 Go 3 action camera, for assessing endurance and maneuverability. The drone camera streamed live footage to a POV headset, offering real-time observation. The SpineWave was wirelessly controlled, with a DJI Mini 3 drone tracking its trajectory. Both the robot and drone, with controllers, are shown in Fig. 6B, and operators in action in Fig. 6C.

Fig. 6 showcases the 50-minute trajectory of the robot in the water ring, covering a roundtrip distance of 800 meters. The robot showed excellent speed, battery life, and a small turning radius, navigating complex natural waters with minimal control latency. Inset footage from the action camera provides the robot's perspective along the ring. For more details, see Supplementary Movie S10.

Notably, during our operation, our robot had two opportunities to interact with the local wildlife, approaching an egret without any disturbance and capturing intimate footage (Fig. 6 D.1, D.2), which highlights the biomimetic design's potential for environmental monitoring.

We also tested the robot in the Deep-Sea Pavilion of Polar Ocean Park, a 6m-deep environment with diverse marine life. This setting presented challenges due to varying scales and behaviors of organisms. The single-jointed SpineWave was adjusted for seawater density, minimizing energy requirements for vertical maneuvers. It was equipped with an LED illuminator and waterproof camera to capture POV footage (Fig. 7D), while a diver and stationary camera tracked its movements.

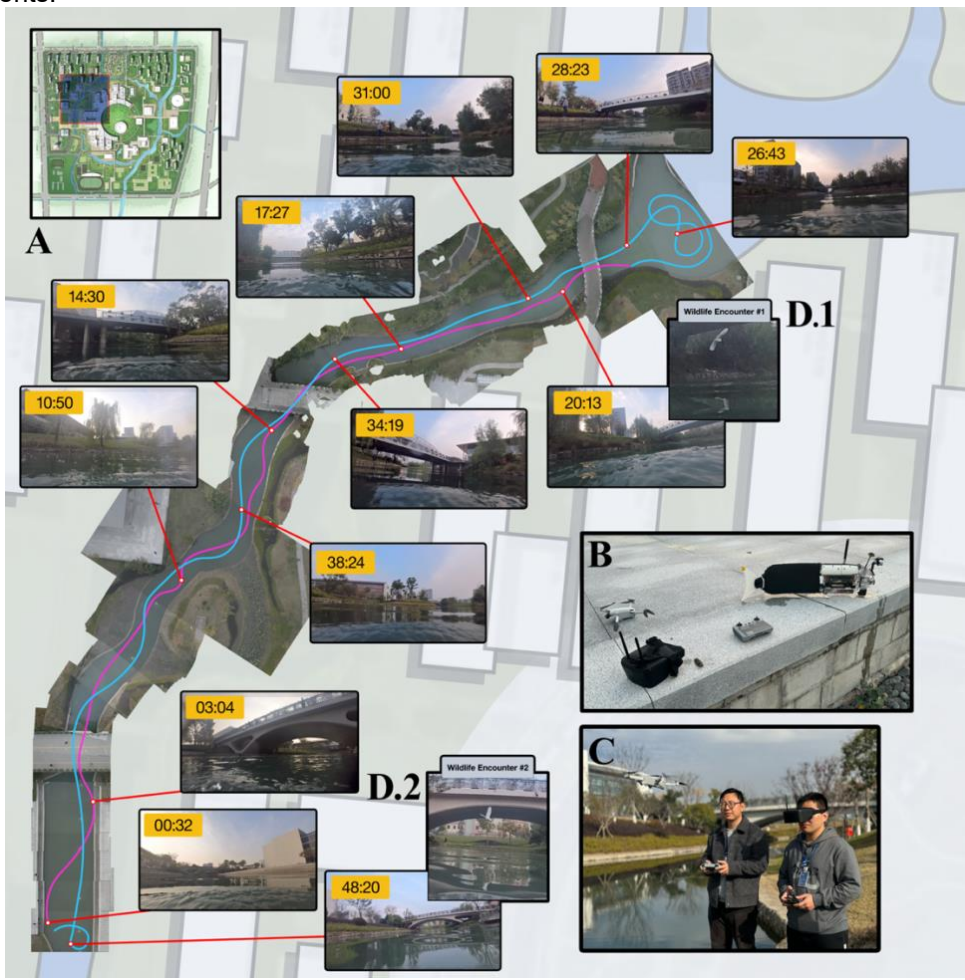

**Fig.6 Overhead drone footage tracking the trajectory of the biomimetic robotic fish, composed onto a 2D map of Westlake University**. (**A**) The sequence includes snapshots from the biomimetic robotic fish's head camera, each annotated with time stamps. (**B**) shows the robotic fish, the streaming POV head-mounted display, and the drone used to monitor the robot from above. In (**C**), a teleoperator controls the robot using a wireless controller as input and the headset for real-time video monitoring. Another operator controls the drone and gives additional feedback to the teleoperator based on the video feed from above. (**D.1** and **D.2**) show two instances when local wildlife was captured on camera, showcasing the capability of our platform.

The single-jointed *SpineWave* showed smooth speed transitions and precise pitch control (Fig. 7A), achieving a speed of 0.296 m/s (0.58 body length per second) and completing a 180-degree turn in 15 seconds. Over eight dives across three days, totaling 200 minutes, it reached a depth of 4.2 meters and swam over 2 km in total.

Preliminary observations revealed the *SpineWave* could interact with marine life without causing disturbance. Even at close range, fish did not flee (Fig. 7B). Fig. 7C shows the *SpineWave* navigating narrow openings. Videos S11-S12 highlight the *SpineWave*'s exploration and interactions, demonstrating its potential for seamless integration into natural underwater environments.

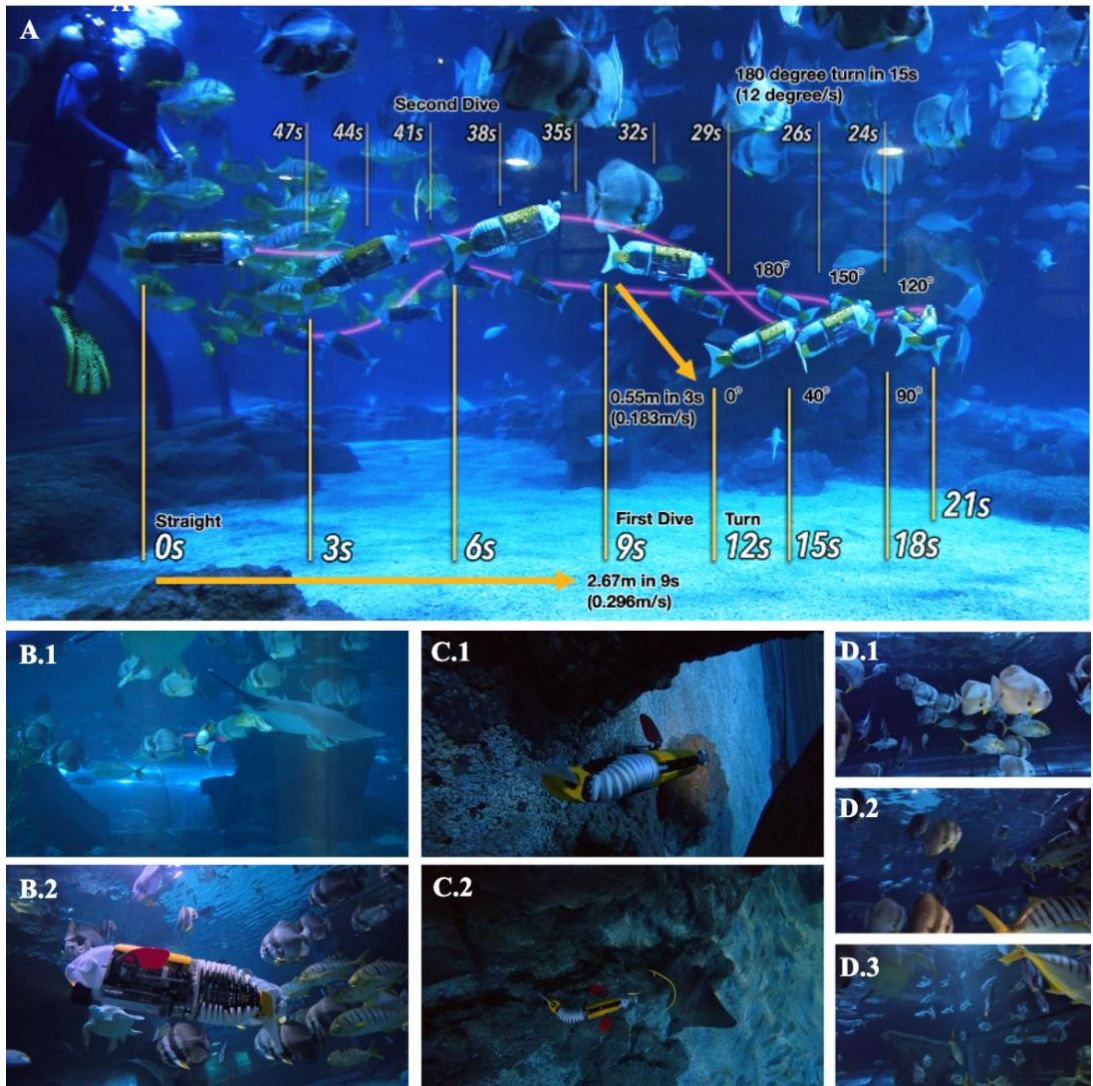

**Fig.7 Swimming in a simulated undersea environment.** (A) *SpineWave*'s trajectory is observed from outside through the aquarium window. *SpineWave*'s trajectory is captured from video footage every 3 seconds and then compiled into a stacked image. (B) *SpineWave* integrates closely with real marine life. (C) *SpineWave* explores underwater environments and narrow openings. (D) *SpineWave*'s head-mounted camera captures the image.

## Discussion

The *SpineWave* bionic fish robot draws inspiration from fish anatomy, combining rigid and flexible structures with magnetic constraints for biomimetic underwater propulsion. By blending the natural flexibility observed in fish with the strength of rigid structures, we have engineered a biomimetic robotic fish capable of exploring in various environments.

The modular design, as the cornerstone of our design, enables swift hardware iteration and on-site adjustments for diverse mission needs, such as close-range observation, water quality testing, and swimming feeding. Three biomimetic prototypes demonstrate versatility, with dedicated fin modules for enhanced swimming capabilities.

To optimize swimming fluid dynamics, we implemented an online evolutionary training algorithm, focusing on straight-line swimming and turning. The EGO algorithm significantly improved the robot's swimming performance with minimal additional real samples. Field trials in open-water environments validated SpineWave's performance, confirming its potential for environmentally friendly robotic tasks[61].

Currently, the modular exoskeleton uses neodymium magnets for uniform stiffness, but research suggests non-uniform stiffness could improve swimming efficiency[62] Future optimization could include active electromagnets, offering tunable stiffness for better performance across environmental conditions and swimming speeds[27,63]. Furthermore, Future designs may incorporate tunable stiffness[64] and distributed actuators like pneumatic muscles[65] or dielectric elastomer actuators (DEA)[66] for enhanced efficiency[67]. Notedly, the magnet implementation has no moving parts, making it suitable for prolonged use in harsh environments like deep water with high pressure[68] or corrosive coastal areas[69].

Additionally, a current limitation involves the relatively long, rigid head segment, necessary for housing core components, which may affect swimming stability due to its lack of active stabilization; while modular buoyancy elements help mitigate this, it represents a trade-off between functionality and optimal biomimetic dynamics.

While our focus is on replicating fish spine function, and we currently use cotton stockings to cover the body. We acknowledge the multifunctional roles of fish skin in flexibility[70], adaptability[71], hydrodynamic efficiency[72], and sensory capabilities[73]. Advancements in bio-inspired sensing technologies[74], especially the flexible sensing[75], will allow future robots to enhance environmental awareness[76,77], dynamically adjust control parameters[78,79], mimicking the adaptive behavior of living fish.

In the current evolutionary learning framework, we primarily use EGO optimization, which adapts to various flow conditions. While EGO is shown to be efficient in automate evolutionary optimization experiments[80] to maximize Expected Improvement (EI), it does come with certain limitations. it has limitations in computational cost as iterations increase[81] and dimensions gets larger[82]. Deep neural networks (DNNs) could manage larger datasets and control parameters, enabling more complex behaviors[83]. Combining this with reinforcement learning techniques[84] will allow the SpineWave to evolve (with advanced sensing and fin actuation system) and adapt in both laboratory and real-world environments[85]. These advancements will bring us closer to designing robotic systems with unparalleled versatility and efficiency, truly reflecting the complex adaptations of biological fish.

## Materials and Methods

In this section, we first document the design and fabrication of electronics, key structural components, and magnetic ribcage. Then, we review the CPG control model and the EGO algorithm.

### Electronics Design and Fabrication

The electronic components of our biomimetic robotic fish are housed in a watertight container made from a transparent acrylic tube and 3D-printed endcaps. This design allows for easy inspection, leakage detection, and visibility of the controller's debugging LCD. A custom PCB (Supplementary Material Note S8) consolidates wiring and provides structural support. To meet the high peak current demands of the servo motors during high-acceleration motion, we powered them directly from an 8.4V 2S Li-ion battery. Two step-down voltage regulators supply stable 5V and 3.3V outputs for the controller and sensors. Communication is handled by serial-to-433 MHz wireless modules, chosen for their signal penetration through water and extended range, enabling parameter exchange and real-time adjustments to the swimming patterns.

### Structural Components Design and Fabrication

We use additive manufacturing methods like Selective Laser Sintering (SLS) and Stereolithography (SLA) for the structural components of the biomimetic robotic fish. Materials were chosen based on mechanical and environmental requirements. SLA resin was used for water-tight parts with lower strength needs, while SLS nylon suited areas with moderate stress. For high-stress sections requiring rigidity, we selected SLS glass fiber–reinforced nylon. SLM titanium alloy was used for critical components like the servo motor bracket, ensuring strength, durability, and chemical resistance.

### Magnetic ribcage design and fabrication

The magnetic ribcage, a key structural component, is made from SLS nylon and consists of vertebrae and magnetic rail units. Each assembly includes two vertebrae and four N52 neodymium magnets (4mm x 5mm) in the rail units. These modular rail units can be replaced with two M2 screws for easy adjustments. The magnets are customizable for different operational needs. When assembled, the ribcage forms a robust, protective frame, enhancing the robot's structural integrity. The vertebrae are connected with rotation joints using self-lubricating brass bushings, locknuts, and M2 screws, all selected for corrosion resistance and maintenance ease in aquatic environments. The ribcage underwent extensive testing, including a 10-day submersion. Its dynamic behavior showed minimal change, confirming its durability. Theoretical models predicting magnetic forces and interactions between the endoskeleton and exoskeleton ribcages are detailed in Supplementary Fig S8.

**Central pattern generators control method**

We implemented the Central Pattern Generator (CPG) control model for our robot. CPGs are neural networks that generate rhythmic motor outputs independently, similar to biological circuits controlling activities like walking and swimming[57]. Our robot uses a CPG model to achieve coordinated movements of the body joints and pectoral fins, ensuring a lifelike motion. This CPG model reduces control complexity significantly, offering simplicity for optimization and practical use. The specific formulation of the CPG used is detailed as follows:

$$\dot{u}_i = hu_{i-1}cos\theta - j(v_{i-1} - b)sin\theta + k^2 u_i(\varepsilon_i - \gamma_i^2) - \omega v_i$$

$$\dot{v}_i = hu_{i+1}sin\theta + j(v_{i+1} - b)cos\theta + k^2 v_i(\varepsilon_i - \gamma_i^2) + \omega u_i \qquad (1)$$

where $u$ and $v$ are the state variables of the oscillators, and $\gamma = \sqrt{u^2 + (v-b)^2}$ ($b$ is the offset vector). $\omega$ and $\varepsilon$ are the intrinsic frequency and amplitude, respectively. Additionally, $k$ is the gain coefficient. $\theta$ donates the phase difference between each oscillator and $i(i > 0)$ means the number of the oscillator. Function $\cdot$ stands for the inner product operation. $h, j$ are the specific coupling coefficients.

**Efficient global optimization**

Efficient global optimization (EGO) is a well-known sequential adaptive sampling methods for black-box optimization problems, and its key steps have been discussed in the evolutionary propulsion optimization section. In EGO, the Kriging surrogate is used to replace expensive black-box functions. The prediction output of a Kriging model can be expressed as a linear combination of the output responses of the training points near the prediction point[86], as follows:

$$\hat{y} = p(x) + Z(x) \qquad (4)$$

where $p(x)$ represents a polynomial that globally approximates the real response. $Z(x)$ is a zero-mean Gaussian process with spatial correlation:

$$COV[Z(x_i), Z(x_j) = \delta^2 R(x_i, x_j)] \qquad (5)$$

where $\delta^2$ indicates the process variance, and $R(x_i, x_j)$ refers to the correlation function between two training points $x_i$ and $x_j$. The Gaussian correlation function typically used is:

$$R(\theta) = \prod_{k=1}^{Dim} \exp(-\theta_k d_k^2) \qquad (6)$$

where $Dim$ indicates the number of decision variables, $d_k$ is the Euclidean distance between $x_i$ and $x_j$, and $\theta_k$ represents the unknown parameter vector to be determined. Then the predicted value $\hat{y}$ at the prediction point $x$ is calculated as:

$$\hat{y} = \hat{\beta} + r^T(x)R^{-1}(y - \hat{\beta}p) \qquad (7)$$

where $r^T(x)$ indicates the correlation vector of length between $x$ and the training points, $y$ is the real responses at the training points. The scalar $\hat{\beta}$ and $r^T(x)$ can be expressed as:

$$\hat{\beta} = \frac{(p^T R^{-1} p)^{-1} p^T R^{-1} y}{N} \tag{8}$$

$$r^T(x) = [R(x, x_1), R(x, x_2), \ldots, R(x, x_N)]^T \tag{9}$$

The variance of the output model is estimated as:

$$\hat{\sigma}^2 = \frac{(y - \hat{\beta} p)^T R^{-1} (y - \hat{\beta} p)}{N} \tag{10}$$

The unknown parameters $\theta_k$ can be estimated by solving a constrained maximization problem:

$$\text{Max: } \Phi(\theta) = \frac{-[N \ln(\hat{\sigma}^2) + \ln|R|]}{2} \tag{11}$$

$$s.t.: \theta > 0 \tag{12}$$

Once the Kriging model is established, we update the next sampling point $x_b$ by finding the maximum of the expected improvement-based (EI) function, as follows:

$$EI(x) = \left(f_{min} - \hat{f}(x)\right) \Phi\left(\frac{f_{min} - \hat{f}(x)}{\hat{\sigma}(x)}\right) + \hat{\sigma}(x) \varphi\left(\frac{f_{min} - \hat{f}(x)}{\hat{\sigma}(x)}\right) \quad s > 0 \tag{13}$$

where $f_{min}$ is the minimum value in the database, $\Phi$ and $\varphi$ are the standard normal cumulative distribution function and the standard normal distribution probability density function, respectively. When $s = 0$, EI value of $x$ is set to 0. The pseudocode code of EGO is shown in Algorithm 1.

---

**Algorithm 1** The framework of EGO.

**Input:** Initial training set $T_r$, Initial population $Pop$, Maximum number of infill points $N_{mf}$, Crossover and mutation probabilities $p_c$, $p_m$ of GA, Real objective function $f$, Parameter vector of the Kriging model $\eta$.
1:   Initialize the database *DB* with the training data set (*LHS* with 10*Dim*).
2:   Kriging modeling $\hat{f}_k(x)$.
3:   $ct = 0$.
4:   **while** $ct < N_{mf}$ **do** (5*Dim*)
5:     Expected improvement-based infill criteria using the GA and obtain the current best solution $x_b$.
6:     Evaluate $x_b$ with the real objective function.
7:     Update *DB* & $\hat{f}_k(x)$ with $x_b$.
8:   **end while**
**Output:** The optimal solution.

**Acknowledgments**

We thank Boai Sun, Fei Han, Xinyu Zeng, Zhen Yang, and Anqi Zhang for their contributions.

**Funding:**

Innovative Research Foundation of Ship General Performance under Grant No. 31422225

the National Key Research and Development Program under Grant No. 2022YFC2805200

the Research Center for Industries of the Future at Westlake University

Westlake Education Foundation under Grant No. WU2022C036

Scientific Research Funding Project of Westlake University under Grant No. 2021WUFP017

**Author contributions:**

W.K. Li. conceptualized the biomimetic robotic fish. Q. He. developed the biomimetic robotic fish, including the design and fabrication of the interface module. W.K. Li. developed the locomotion model and control software of the biomimetic robotic fish. G. M. Dai developed the simulation of bionic exoskeletons and data processing of experiments. H. Chen developed the evolutionary algorithm in experiments. Q. M. Liu and J. You developed the algorithms for collecting and integrating all data sensors in the experiment. Q. He., G. M. Dai, and W.K. Li. performed pool experiments as well as water ring experiments and wrote the paper. X.Q. Tian was responsible for circulating water tank and optimization experiments. W. C. Cui, M.S. Triantafyllou and D. X. Fan were responsible for the overall research direction, objectives, and funding.

**Competing interests:** The authors declare that they have no competing financial interests.

**Data and materials availability:** All data needed to evaluate the conclusions in the paper are present in the paper or the Supplementary Materials.